\documentclass[conference]{IEEEtran}
\IEEEoverridecommandlockouts
\usepackage{cite}
\usepackage{amsmath,amssymb,amsfonts}
\usepackage{algorithm,algorithmic}
\usepackage{graphicx}
\usepackage{textcomp}
\usepackage{float}
\usepackage{caption}
\usepackage{subcaption}
\usepackage{xcolor}
\usepackage{makecell}
\usepackage{multirow}
\usepackage{lipsum}
\usepackage{hyperref}
\usepackage{adjustbox}
\usepackage{afterpage}
\usepackage{graphicx}
\usepackage{subcaption}  

\def\BibTeX{{\rm B\kern-.05em{\sc i\kern-.025em b}\kern-.08em
    T\kern-.1667em\lower.7ex\hbox{E}\kern-.125emX}}
    
\begin{document}

\title{Search-Based Robot Motion Planning With Distance-Based Adaptive Motion Primitives\\
}



\author{
\IEEEauthorblockN{
Benjamin Kraljušić\textsuperscript{1},
Zlatan Ajanović\textsuperscript{2},
Nermin Čović\textsuperscript{1},
Bakir Lačević\textsuperscript{1}
}
\IEEEauthorblockA{
\textsuperscript{1}Faculty of Electrical Engineering, University of Sarajevo, Bosnia and Herzegovina \\
\textsuperscript{2}RWTH Aachen University, Germany \\
\{bkraljusic1, ncovic1, blacevic1\}@etf.unsa.ba\textsuperscript{1}, zlatan.ajanovic@ml.rwth-aachen.de\textsuperscript{2}
}
}


\maketitle

\begin{abstract}
This work proposes a motion planning algorithm for robotic manipulators that combines sampling-based and search-based planning methods. The core contribution of the proposed approach is the usage of burs of free configuration space ($\mathcal{C}$-space) as adaptive motion primitives within the graph search algorithm. Due to their feature to adaptively expand in free $\mathcal{C}$-space, burs enable more efficient exploration of the configuration space compared to fixed-sized motion primitives, significantly reducing the time to find a valid path and the number of required expansions. The algorithm is implemented within the existing SMPL (Search-Based Motion Planning Library) library and evaluated through a series of different scenarios involving manipulators with varying number of degrees-of-freedom (DoF) and environment complexity. Results demonstrate that the bur-based approach outperforms fixed-primitive planning in complex scenarios, particularly for high DoF manipulators, while achieving comparable performance in simpler scenarios. 
\end{abstract}


\section{Introduction}
\indent Sampling-based motion planning algorithms aim to avoid the explicit construction of $\mathcal{C}$-space~\cite{LaVallePlanningAlgorithms}. The core idea of these algorithms is to generate a set of configurations by sampling the configuration space and attempting to locally connect them using collision-checking routines. This process yields a valid sequence of configurations -- a path between the start and goal configurations of a robotic manipulator~\cite{LacevicBurs}.
\\ \indent On the other hand, search-based motion planning algorithms treat the motion planning problem as a graph search problem. Due to the discrete nature of graphs, all search-based planning algorithms require a discrete representation of the $\mathcal{C}$-space either by sampling using motion primitives, state lattice or  some other discretization method. Thus, the planning problem is reduced to constructing an appropriate graph representation of the configuration space and performing a search over this graph to connect the start with the goal configuration~\cite{SotirchosSMPLvsOMPL}.
\\ \indent In this work, a motion planning algorithm is presented that performs heuristic graph search over a graph generated by sampling-based methods, with the aim of efficiently finding a solution that may be optimal or within a bounded level of suboptimality. Local paths within the graph are generated using burs of free configuration space -- a planning structure proposed in~\cite{LacevicBurs}. Burs were chosen due to their proven exploration capabilities, as shown in the existing work. The ARA$^{\ast}$ algorithm~\cite{LikhachevARAStar} was used as the search algorithm, as it quickly finds an initial solution and then efficiently improves it through an iterative process, relying on the results of previous searches. The main contribution of this paper represents developing an algorithm that carefully combines the advantages of both planning paradigms to efficiently find good paths.

\begin{figure}[!t]
    \centering
    \includegraphics[scale=0.18]{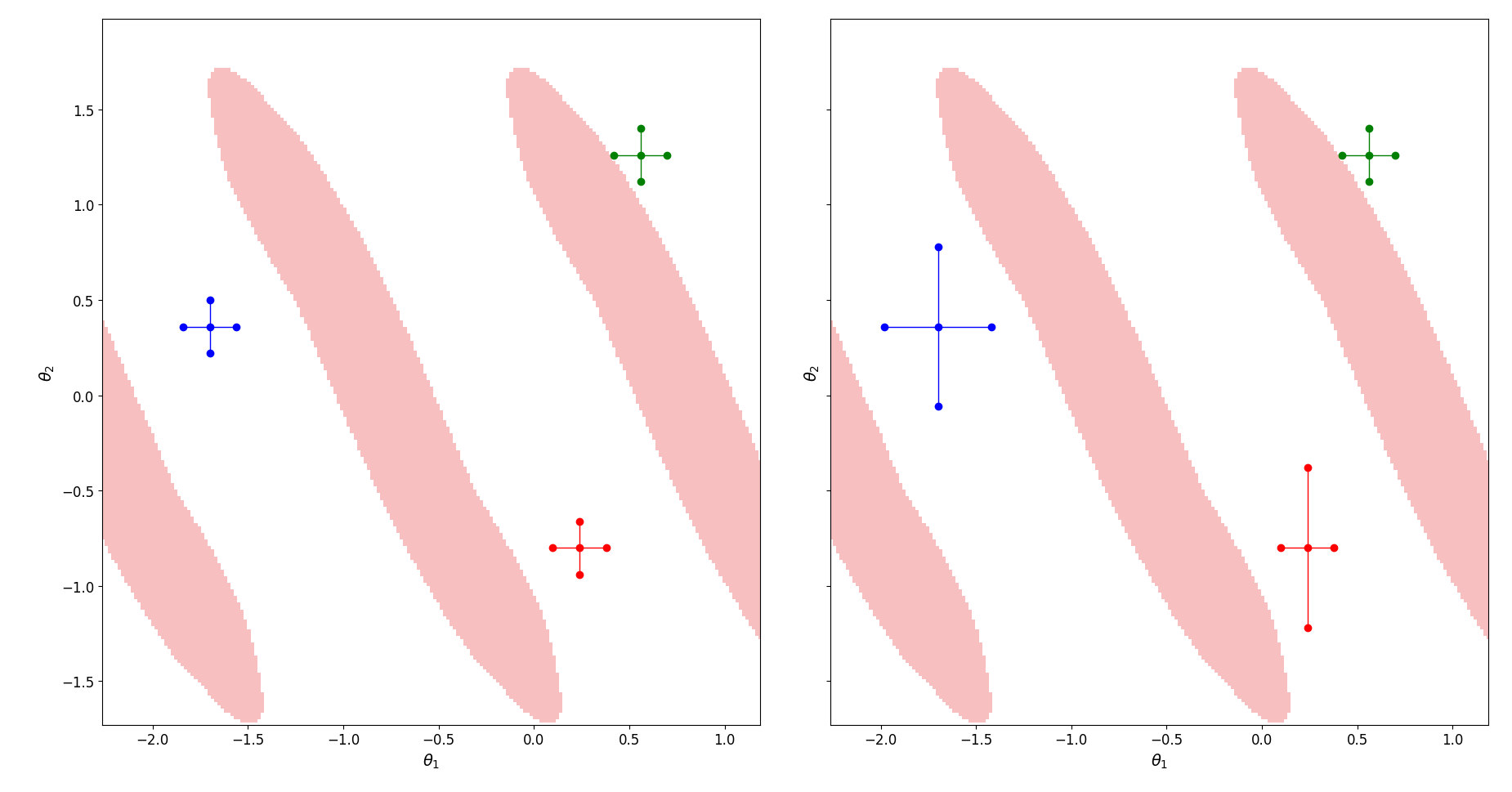}
    \caption{Expansion of different states using fixed motion primitives (left) and adaptive bur-based primitives (right) for scenario shown in the figure \ref{fig:2DoFMed}. Pink-colored areas represent the $\mathcal{C}$-space obstacles.}
    \label{fig:burVsMprim}
\end{figure}

\section{Related Work}
Sampling-based planning algorithms are most commonly divided into two categories: Rapidly-Exploring Random Trees (RRT)~\cite{LaValleRRT} and Probabilistic Roadmaps (PRM)~\cite{KavrakiPRM}. The PRM algorithm constructs a roadmap in the form of a graph, where nodes represent collision-free configurations of the robotic manipulator, and edges correspond to valid paths between those nodes. The initial and goal configurations are connected (if possible) to any two nodes in the roadmap, and a path between them is found via graph search~\cite{KavrakiPRM}. On the other hand, RRT algorithm builds a tree from the initial configuration by incrementally connecting randomly sampled configurations using collision-checking routines. The algorithm terminates when the goal configuration is added to the tree. Many enhanced versions of these basic algorithms have been proposed. A more efficient variant of RRT is its bidirectional version, RRT-Connect~\cite{KuffnerRRTConnect}, which constructs two trees -- one from the initial and the other from the goal configuration and attempts to connect them. Asymptotically optimal versions of these algorithms, RRT$^{\ast}$ and PRM$^{\ast}$, have also been introduced~\cite{KaramanOptimalneVerzije}. Orthey et al. provide a comprehensive overview and comparative review of sampling-based motion planning algorithms~\cite{SamplingBasedPlanningReview}.
\\ \indent One of the most notable graph search algorithms used for motion planning of robotic manipulators is A$^{\ast}$, developed in 1968 during the \textit{Shakey} project~\cite{AStarSearch}. The A$^{\ast}$ algorithm, considered a generalization of Dijkstra’s shortest path algorithm~\cite{Dijkstra}, uses a heuristic function to estimate the cost to the goal, thereby accelerating the search process. Although these algorithms are complete and find optimal solutions, their efficiency decreases significantly with increasing dimensionality of the search space. Higher dimensionality causes an exponential growth in the number of graph nodes (a so-called “curse of dimensionality”)~\cite{Bellman}. Various variants of the A$^{\ast}$ algorithm have been developed to improve performance under different planning scenarios. In general, performance, such as execution time and memory usage, can be improved by introducing weighting coefficients to the heuristic function, allowing for limited suboptimality of the solution~\cite{WeightedAStarUnify}. The idea of the Weighted A$^{\ast}$ algorithm was first introduced by Pohl~\cite{WeightedAStar}, who later proposed dynamic weighting coefficients~\cite{DWeightedAStar}. Based on the observation that planning time is often limited in real-world problems, Likhachev et al. developed Anytime Repairing A$^{\ast}$ (ARA$^{\ast}$), which uses weighted heuristic search to quickly find a bounded-suboptimal solution and then refines it within the remaining time to reduce suboptimality~\cite{LikhachevARAStar}. Aine et al. showed that combining multiple heuristic functions can guide the search more effectively, resulting in faster discovery of acceptable-quality solutions~\cite{MultiHeuristicAStar}. An alternative approach is presented in~\cite{MultiResolutionAStar} and ~\cite{IncrementalAdaptive}, where search acceleration is achieved using graphs of different resolutions.
\\ \indent Pivtoraiko and Kelly introduced the concept of a state lattice. A state lattice is a graph representation of the configuration space where nodes represent configurations and edges correspond to feasible transitions between neighboring configurations~\cite{PivotraikoLattice}. These feasible transitions respect the system’s physical constraints. State lattices have proven to be a useful method for representing configuration spaces when searching for dynamically feasible trajectories~\cite{LikhachevLong}.
\\ \indent Cohen et al. developed a motion planner that performs heuristic search over the configuration space and constructs a graph using motion primitives—minimal executable motions of the robotic manipulator~\cite{CohenFixedMotionPrimitives}. This planner showed substantial performance in high-dimensional configuration spaces. In a subsequent paper, the authors extended the algorithm by introducing adaptive motion primitives and enabling planning in a lower-dimensional space when possible~\cite{CohenAdaptiveMotionPrimitives}. A different form of adaptive motion primitives, applicable to manipulators with single or dual robotic arms, is presented in~\cite{PrimitivesSingleArmDualArm}. Adaptive primitives for automated parking are presented in~\cite{AdabalaICAT}. However, these primitives are rule-based and can not be generalized to different $\mathcal{C}$ - spaces.
\\ \indent Sotirchos and Ajanović provide a detailed comparison between sampling-based and search-based motion planning algorithms. In~\cite{SotirchosSMPLvsOMPL}, they compare the performance of the RRT-Connect and ARA$^{\ast}$ algorithms as prominent representatives of both categories. They show that sampling-based algorithms generally provide more consistent performance across planning scenarios, whereas the performance of search-based algorithms can be significantly improved by tailoring the graph and search method to the specific problem.
\\ \indent The concept of bubbles of free configuration space is firstly introduced by Quinlan. A bubble is a local volume around a given configuration that is guaranteedly free of collisions~\cite{QuinlanBubbles}. Its computation is based on a single distance information between the robot and the nearest obstacle in the workspace. Several studies have subsequently investigated motion planning using such distance information. In~\cite{LacevicBubblesAStar} and~\cite{LacevicBubblesAStar2}, the authors proposed a path planning method for constructing a collision-free tree using bubbles. In~\cite{AdemovicLacevicEvolucija}, an improved algorithm is presented using such bubble-based tree structure and a heuristic-guided search combined with an evolutionary learning algorithm, yielding a better exploration of the configuration space.
\\ \indent The computation of bubbles of free $\mathcal{C}$-space relies on conservative assumptions to maintain convexity and computational efficiency~\cite{LacevicBurs, QuinlanBubbles}. Lačević et al.~\cite{LacevicBurs} introduced a new planning structure called the \textit{bur of free C-space}. Although the computation of a bur uses the same distance information as bubbles, its boundary extends significantly beyond that of a free-space bubble. In the same work, they integrated this structure into a tree-based planner (bur tree) based on RRT-Connect~\cite{KuffnerRRTConnect}. The proposed RBT-Connect algorithm demonstrated remarkable reduction of planning times, number of iterations, and node counts. The authors further extended the algorithm to rigid bodies moving in 2D and 3D configuration spaces~\cite{LacevicBursRigid}. Lačević et al. later introduced the concept of the \textit{generalized bur}~\cite{LacevicGBurs}, which significantly enlarges the original bur using both distance to obstacles and its corresponding underestimation. The proposed structure provided RGBT-Connect algorithm, which conquered RBT-Connect according to all criteria used for the comparison. Čović et al. developed an asymptotically optimal version of the algorithm, RGBMT$^{\ast}$~\cite{CovicMulti}, which builds generalized bur trees from configurations beyond just the initial and goal states, thus increasing the $\mathcal{C}$-space exploration even more. These trees are then optimally connected to yield an optimal path to the goal. While all these algorithms assume static environments, Čović et al. also proposed a method which exploits generalized burs for motion planning in dynamic environments~\cite{CovicDinamika}. 
\\ \indent
Our work builds upon existing research by introducing \textit{burs of free} $\mathcal{C}$-space as adaptive motion primitives, which are utilized to construct a search graph, thereby enhancing existing search-based planning methods.
\\ \indent This work is organized as follows. Sec. III presents and elaborates on the proposed motion planning approach. Sec. IV presents the simulation study. Simulation setup and evaluation metrics are explained, and the simulation results are presented. Finally, Sec. V brings some conclusion remarks and future work directions.

\section{The Proposed Motion Planning Approach}
This section presents the motion planning algorithm proposed in this work, which performs graph search over a structure constructed using adaptive motion primitives, based on burs of free configuration space. We begin by describing the proposed approach in detail, followed by an explanation of the existing planning solution implemented in SMPL. Finally, we outline how our method has been integrated into the same framework.

\subsection{Graph Construction and Motion Primitives}
The graph is usually built incrementally during search as nodes are expanded and successor nodes generated using motion primitives, avoiding the need to store a full high-dimensional graph. 
Often, neighboring successor states are generated using static motion primitives~\cite{CohenFixedMotionPrimitives,CohenAdaptiveMotionPrimitives}, where each primitive represents the smallest unit of motion for a single joint. The approach proposed in \cite{CohenFixedMotionPrimitives} employs fixed primitives, allowing each $i$-th joint, $i\in{1,2,\dots,n}$, where $n$ is the number of DoFs, to move by a fixed angle $\theta_i$ in both directions. For a manipulator with $n$ DoFs, $2n$ motion primitives are defined. Additionally, in every expansion step, the algorithm attempts a direct connection to the goal, enabling planning to succeed even if the goal is not part of the graph.

Graph resolution is determined by the motion primitive length. Larger primitive steps reduce the number of nodes and potentially the search time, but can hinder completeness and optimality, especially in environments with narrow passages. Smaller steps provide finer resolution and better path quality at the cost of increased search time~\cite{MultiResolutionAStar}.

\subsection{Burs-based Adaptive Motion Primitives}

Burs are well-suited for adaptive motion primitives generation, as they provide provable collision-free spines (primitives) connecting the center configuration (initial state) to many reachable states while maximizing the step for each primitive~\cite{LacevicBurs}. 

Bur construction requires information of the minimum distance between the robot and obstacles. In this work, to facilitate collision/distance queries, voxel-based workspace modeling and a sphere-tree robot model~\cite{OsullivanSpheres} are used. The distance is calculated between leaf spheres and the closest occupied voxel. While using leaf spheres offers accuracy, it is computationally expensive. Larger spheres from higher tree levels offer a conservative but cheaper alternative since there is not so many of them compared to leaf spheres. However, we chose accurate distance estimation using leaf spheres.

To match the fixed-primitive search structure and clearer comparison to the baseline, bur spines correspond to single-joint motions. 

When $d_c$ is small, bur spines become short. If $d_c < d_{crit}$, where $d_{crit}$ is a suitably user-defined parameter (0.03 [m] in this work), or the spine is shorter than the primitive length, neighbors are generated using fixed primitives instead. Thus, in cluttered areas, burs degrade to fixed primitive structures, similarly as RGBT algorithm switches to RRT-mode (see ~\cite{LacevicBurs,LacevicGBurs}). Both structures that capture collision free area of configuration space are shown in Figure~\ref{fig:burVsMprim}.

Bur spines are discretized by rounding their length to the nearest and lowest integer multiple of the primitive length, (i.e., using the floor function $\lfloor\cdot\rfloor$), allowing consistent graph discretization for both approaches.


\subsection{Graph Search Algorithm}
Any graph search algorithm can be applied to a bur-tree. This work adapts the ARA$^\ast$ algorithm described previously. The algorithm we propose is presented in algorithms \ref{alg:proposed} and \ref{alg:improve}.

\begin{algorithm}
{\scriptsize
    \caption{ARA$^{\ast}$ search with burs}
    \begin{algorithmic}[1]
        \renewcommand{\algorithmicrequire}{\textbf{Input:}}
        \renewcommand{\algorithmicensure}{\textbf{Output:}}
        \REQUIRE{\phantom{j}\\
        $s_{start}$: start node,\\
        $s_{goal}$: goal node,\\
        $\varepsilon > 1$}: suboptimality bound,\\
        $t_{allowed}$: allowed planning time \\    
        \ENSURE{\phantom{j}\\$(sub)optimal\_path$ from $s_{start}$ to $s_{goal}$\\\phantom{j}} \\
        
\STATE $g(s_{goal}) \leftarrow \infty$, $g(s_{start}) \leftarrow 0$ \\
\STATE $open\_set \leftarrow \{start\}$ \\
\STATE $closed\_set \leftarrow \emptyset$, $incons\_set \leftarrow \emptyset$ \\
\STATE $f(s_{start}) \leftarrow g(s_{start}) + \varepsilon \cdot h(s_{start})$ \\
\STATE \textbf{call} \texttt{ImprovePathUsingBurs()} \\
\STATE $\varepsilon' \leftarrow \min\left\{\varepsilon, \frac{g(s_{goal})}{\min_{s \in open\_set \cup incons\_set}(g(s)+h(s))}\right\}$ \\
\WHILE{\texttt{getElapsedTime()} $< t_{allowed}$} { \label{elaps}
\STATE save the latest $\varepsilon'$-suboptimal solution \\

\WHILE{$\varepsilon' > 1$}{
    \STATE $\varepsilon \leftarrow \varepsilon - \Delta\varepsilon$ \\
    \STATE move nodes from $incons\_set$ to $open\_set$ \\
    \STATE sort $open\_set$ based on $f(s)$ \\
    \STATE $closed\_set \leftarrow \emptyset$ \\
    \STATE \textbf{call} \texttt{improvePathUsingBurs()} \\
    \STATE $\varepsilon' \leftarrow \min\left\{\varepsilon, \frac{g(s_{goal})}{\min_{s \in open\_set \cup incons\_set}(g(s)+h(s))}\right\}$ \\
    \STATE save the latest $\varepsilon'$-suboptimal solution \\
}\ENDWHILE
\RETURN $(sub)optimal\_path$
}\ENDWHILE
\RETURN $(sub)optimal\_path$
    \end{algorithmic} 
    \label{alg:proposed}
}
\end{algorithm}

        
\begin{algorithm}
{\scriptsize
    \caption{\texttt{improvePathUsingBurs()}}
    \begin{algorithmic}[1]
    \WHILE{$f(s_\text{goal}) > \min_{s \in \text{open\_set}}(f(s))$}
        \STATE remove $s$ with smallest $f(s)$ from $open\_set$
        \STATE add $s$ to $closed\_set$
        \STATE $d_c \leftarrow \texttt{getDistanceInformation}(\mathbf{q})$
        \STATE $S \leftarrow Bur(\mathbf{q}, Q_e, d_c)$
        \FORALL{$s'$ in $S$}
            \IF{$s'$ not visited}
                \STATE $g(s') \leftarrow \infty$
            \ENDIF 
            \IF{$g(s') > g(s) + c(s, s')$}
                \STATE $g(s') \leftarrow g(s) + c(s, s')$
                \IF{$s' \notin \text{closed\_set}$}
                    \STATE add $s'$ with $f(s')$ to $open\_set$
                \ENDIF
            \ELSE
                \STATE add $s'$ to $incons\_set$
            \ENDIF
        \ENDFOR
    \ENDWHILE
    \end{algorithmic} 
    \label{alg:improve} 
}
\end{algorithm}
Similar to the ARA$^{\ast}$ algorithm~\cite{LikhachevARAStar}, the proposed approach executes a series of A$^{\ast}$ searches, progressively refining the solution obtained in previous iterations. This process continues until either an optimal path is found or the elapsed time exceeds the user-defined planning time limit. The elapsed time is measured using the \texttt{getElapsedTime()} method (line~\ref{elaps}). The core component of the proposed algorithm is the \texttt{improvePathUsingBurs()} method, presented in Algorithm~\ref{alg:improve}, which generates new search nodes by constructing burs of free $\mathcal{C}$-space at configurations corresponding to the nodes selected for expanding by the search algorithm.
The remaining routines follow the existing SMPL implementation and are therefore not discussed in detail here.
\\ \indent When using fixed motion primitives, all edge costs are equal. For burs, edge costs vary by spine length. The cost function $g(n)$ is updated to reflect actual Euclidean distances between neighboring states, but remains consistent for fixed-length primitives. 
\\ \indent
The heuristic $h(n)$ is defined as the Euclidean distance between the current configuration $\boldsymbol{q}$ and the goal configuration $\boldsymbol{q}_{\text{goal}}$.

\section{Simulation Study}
The proposed algorithm was implemented within the SMPL\footnote{https://github.com/aurone/smpl} library and evaluated by comparing its performance to the existing planning solution provided by the same framework\footnote{The code is available online at \\ \href{https://github.com/benjaminkraljusic/bur_search_motion_planning}{github.com/benjaminkraljusic/bur\_search\_motion\_planning}.}. This section presents the simulation results obtained by the comparative analysis.

\begin{figure*}[!t]
\centering
\begin{subfigure}{0.155\textwidth}
  \includegraphics[width=\linewidth]{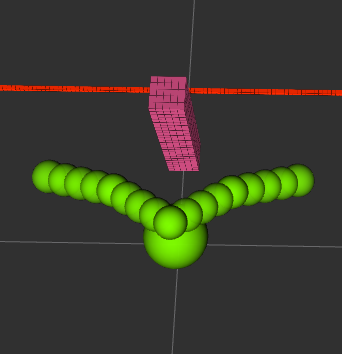}
  \caption{\texttt{2DoF\_EASY}}
  \label{fig:2DoFEasy}
\end{subfigure}\hspace{0.5em}%
\begin{subfigure}{0.155\textwidth}
  \includegraphics[width=\linewidth]{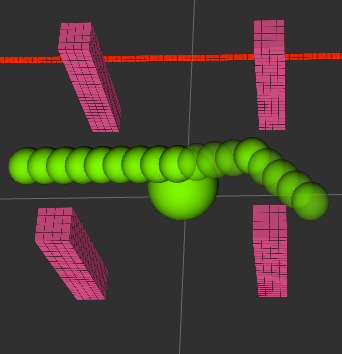}
  \caption{\texttt{2DoF\_MEDIUM}}
  \label{fig:2DoFMed}
\end{subfigure}\hspace{0.5em}%
\begin{subfigure}{0.155\textwidth}
  \includegraphics[width=\linewidth]{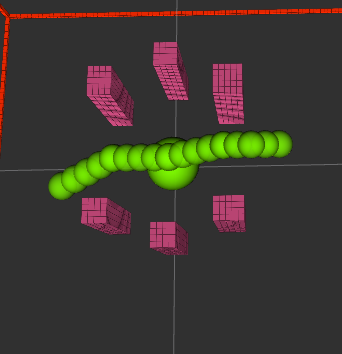}
  \caption{\texttt{2DoF\_HARD}}
\end{subfigure}\hspace{0.5em}%
\begin{subfigure}{0.155\textwidth}
  \includegraphics[width=\linewidth]{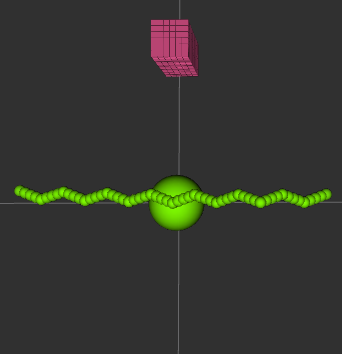}
  \caption{\texttt{7DoF\_EASY}}
\end{subfigure}\hspace{0.5em}%
\begin{subfigure}{0.155\textwidth}
  \includegraphics[width=\linewidth]{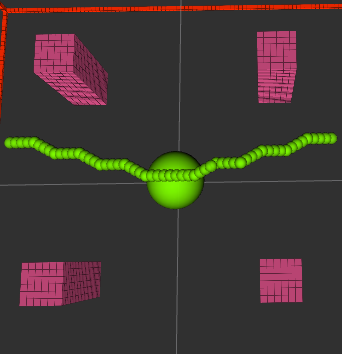}
  \caption{\texttt{7DoF\_MEDIUM}}
\end{subfigure}\hspace{0.5em}%
\begin{subfigure}{0.155\textwidth}
  \includegraphics[width=\linewidth]{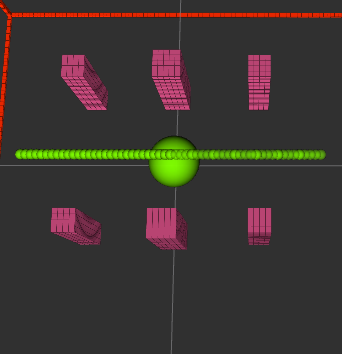}
  \caption{\texttt{7DoF\_HARD}}
  \label{fig:7DoFHard}
\end{subfigure}

\caption{Start and goal configurations for all six planning scenarios used in the simulation study.}
\label{fig:all_scenarios}
\end{figure*}

The simulations are evaluated using two robotic manipulators: a planar two-joint manipulator (\texttt{2DoF}) and a planar seven-joint manipulator (\texttt{7DoF})\footnote{All experiments were run on an \textit{Intel Core i5-8250U} processor with 8 cores at 1.6 GHz.}. Planning with two and seven degrees of freedom examines the algorithm performance with respect to configuration space dimensionality, which is critical for search-based methods.

Three planning scenarios were designed for both manipulators: easy (\texttt{EASY}), medium (\texttt{MEDIUM}), and hard (\texttt{HARD}). Scenario difficulty is based on the number and proximity of obstacles. Easy scenarios contain few obstacles placed far from the manipulator, while harder scenarios consist of many densely packed obstacles. Start and goal configurations for each scenario are shown in Figures~\ref{fig:2DoFEasy}--\ref{fig:7DoFHard}.

The simulations were conducted at various graph resolutions determined by the motion primitive length. The predefined primitive length represents the smallest possible displacement in the configuration space, i.e., the shortest edge length in the search graph. Primitive lengths are ranged from 4 to 12 degrees.

\begin{table}[ht]
\centering
\caption{ARA$^\ast$ parameters for different manipulator types}
\label{tab:parametri-manipulatora}
\begin{tabular}{|c|c|c|c|}
\hline
\textbf{Manipulator Type} & \boldmath$\varepsilon$ & \textbf{Planning Time [s]} & \textbf{Repair Time [s]} \\
\hline
\texttt{2\_DoF}  & 10 & 5  & 1  \\
\texttt{7\_DoF}  & 50 & 60 & 40 \\
\hline
\end{tabular}
\end{table}
Parameters for the ARA$^\ast$ algorithm for both manipulators are given in Table~\ref{tab:parametri-manipulatora}. The parameter $\varepsilon$ defines the initial suboptimality bound. Planning time is the maximum allowed time to find the first solution, which should be a path no more than $\varepsilon$ times longer than the optimal one. Repair time is the additional time allowed to improve the initial solution toward an optimal one.

Performance metrics were selected following the original works proposing search-based planning with motion primitives~\cite{CohenFixedMotionPrimitives, CohenAdaptiveMotionPrimitives}. These include: initial planning time and number of expansions ($t_\text{init}$, $n_\text{init}$), final planning time and expansions ($t_\text{final}$, $n_\text{final}$), and path length ($c$). For scenarios where an optimal solution was not found within the allowed time limits, final planning time and expansions are not reported. Additionally, high-dimensional and high-resolution cases where no solution was found in time are excluded from the results. To mitigate the effect of other background processes, each experiment was repeated 100 times, and the average values for execution time and number of expansions were reported.

\subsection{Results}

\begin{table}[!t]

\scriptsize
\centering
\caption{Planning metrics comparison for scenario \texttt{2DoF\_EASY} (Fixed-length primitives / Burs)}
\label{prvatabela}
\resizebox{\columnwidth}{!}{
\begin{tabular}{|c||c|c|c|c|c|}
\hline
$m_{\text{prim}}$ & $t_{\text{init}}$ [ms] & $n_{\text{init}}$ & $t_{\text{final}}$ [ms] & $n_{\text{final}}$ & $c$ \\
\hline
4  & 6.67 / \textbf{6.39} & 614 / \textbf{578} & 12.63 / \textbf{9.80}  & 1224 / \textbf{914}  & 6.48 / 6.48 \\
5  & 4.33 / \textbf{4.26} & 395 / \textbf{374} & 8.15 / \textbf{6.59}   & 777 / \textbf{600}   & 6.46 / 6.46 \\
6  & 3.61 / \textbf{3.42} & 296 / \textbf{280} & 6.16 / \textbf{5.37}   & 526 / \textbf{464}   & 6.58 / 6.58 \\
7  & 2.98 / \textbf{2.96} & 229 / \textbf{225} & 5.14 / \textbf{3.91}   & 414 / \textbf{308}   & 6.59 / 6.59 \\
8  & 2.43 / \textbf{2.19} & 175 / \textbf{161} & 3.95 / \textbf{3.11}   & 300 / \textbf{240}   & 6.62 / 6.62 \\
9  & 2.15 / \textbf{2.12} & 155 / \textbf{153} & 3.14 / \textbf{2.87}   & 233 / \textbf{216}   & 6.74 / 6.74 \\
10 & \textbf{1.53} / 1.54 & 105 / 105          & 2.50 / \textbf{2.31}   & 184 / \textbf{169}   & 6.55 / 6.55 \\
11 & \textbf{1.60} / 1.63 & 108 / 108          & 2.22 / \textbf{2.08}   & 154 / \textbf{144}   & 6.75 / 6.75 \\
12 & 1.37 / \textbf{1.34} & 90 / \textbf{89}   & 2.05 / \textbf{1.88}   & 138 / \textbf{132}   & 6.91 / 6.91 \\
\hline
\end{tabular}
}
\end{table}

\begin{table}[!t]
\scriptsize
\centering
\caption{Planning metrics comparison for scenario \texttt{2DoF\_MEDIUM} (Fixed-length primitives / Burs)}
\label{drugatabela}
\resizebox{\columnwidth}{!}{
\begin{tabular}{|c||c|c|c|c|c|}
\hline
$m_{\text{prim}}$ & $t_{\text{init}}$ [ms] & $n_{\text{init}}$ & $t_{\text{final}}$ [ms] & $n_{\text{final}}$ & $c$ \\
\hline
4  & 11.08 / \textbf{10.54} & 1084 / \textbf{1009} & 31.97 / \textbf{29.69} & 3316 / \textbf{3016} & 6.85 / 6.85 \\
5  & 7.52 / \textbf{7.09}   & 712 / \textbf{659}   & 19.41 / \textbf{19.23} & 1941 / \textbf{1902} & 6.90 / 6.90 \\
6  & 5.89 / \textbf{5.54}   & 500 / \textbf{485}   & 15.98 / \textbf{14.70} & 1467 / \textbf{1385} & 6.91 / 6.91 \\
7  & 4.41 / \textbf{4.32}   & 350 / \textbf{338}   & 11.94 / \textbf{11.19} & 1009 / \textbf{943}  & 6.93 / 6.93 \\
8  & 3.62 / \textbf{3.56}   & 289 / \textbf{280}   & 9.21 / \textbf{8.87}   & 766 / \textbf{746}   & 7.09 / 7.09 \\
9  & 2.99 / 2.99   & 228 / 228            & 7.10 / \textbf{7.02}   & \textbf{573} / 576   & 6.97 / 6.97 \\
10 & \textbf{2.76} / 2.94   & \textbf{211} / 212   & \textbf{5.80} / 6.05   & \textbf{459} / 460   & 7.23 / 7.23 \\
11 & \textbf{2.25} / 2.48   & \textbf{160} / 162   & \textbf{5.62} / 5.99   & 417 / \textbf{414}   & 7.41 / 7.41 \\
12 & \textbf{2.03} / 2.33   & 145 / 145            & \textbf{4.56} / 5.05   & 340 / \textbf{339}   & 7.09 / 7.09 \\
\hline
\end{tabular}
}
\end{table}

\begin{table}[!t]
\scriptsize
\centering
\caption{Planning metrics comparison for scenario \texttt{2DoF\_HARD} (Fixed-length primitives / Burs)}
\label{zadnjatabela}
\resizebox{\columnwidth}{!}{
\begin{tabular}{|c||c|c|c|c|c|}
\hline
$m_{\text{prim}}$ & $t_{\text{init}}$ [ms] & $n_{\text{init}}$ & $t_{\text{final}}$ [ms] & $n_{\text{final}}$ & $c$ \\
\hline
4  & 12.33 / \textbf{11.99} & 1213 / \textbf{1205} & 13.64 / \textbf{12.67} & 1367 / \textbf{1284} & 7.36 / 7.36 \\
5  & \textbf{7.60} / 7.62   & 746 / \textbf{744}   & \textbf{7.78} / 7.84   & \textbf{764} / 767   & 7.30 / 7.30 \\
6  & 6.11 / \textbf{6.02}   & 535 / 535            & 6.20 / \textbf{6.13}   & \textbf{543} / 545   & 7.37 / 7.37 \\
7  & \textbf{5.17} / 5.23   & 425 / 425            & 5.30 / \textbf{5.29}   & 436 / \textbf{430}   & 7.51 / 7.51 \\
8  & \textbf{3.97} / 3.99   & 322 / 322            & \textbf{4.01} / 4.03   & 325 / \textbf{324}   & 7.52 / 7.52 \\
9  & \textbf{3.23} / 3.32   & 257 / 257            & \textbf{3.24} / 3.33   & 257 / 257            & 7.52 / 7.52 \\
10 & \textbf{2.43} / 2.63   & 198 / 198            & \textbf{2.44} / 2.63   & 198 / 198            & 7.32 / 7.32 \\
11 & \textbf{2.47} / 2.59   & 183 / 183            & \textbf{2.47} / 2.60   & 183 / 183            & 8.19 / 8.19 \\
12 & \textbf{2.07} / 2.13   & 150 / 150            & \textbf{2.07} / 2.14   & 150 / 150            & 7.40 / 7.40 \\
\hline
\end{tabular}
}
\end{table}

\begin{figure*}[!t]
\centering

\begin{subfigure}[t]{0.24\textwidth}
    \centering
    \includegraphics[width=\linewidth]{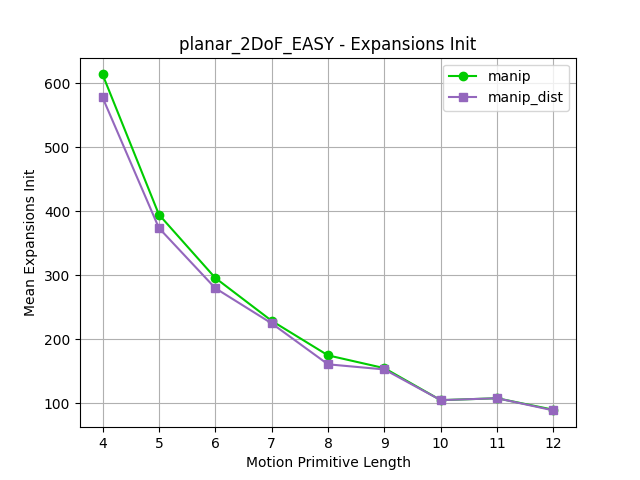}
    \caption{\texttt{2DoF\_EASY} – Expansions}
    \label{fig:prviGrafik}
\end{subfigure}
\hfill
\begin{subfigure}[t]{0.24\textwidth}
    \centering
    \includegraphics[width=\linewidth]{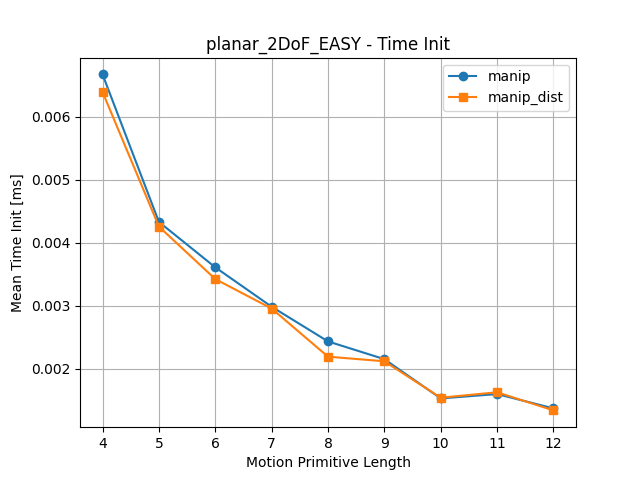}
    \caption{\texttt{2DoF\_EASY} – Planning Time}
\end{subfigure}
\hfill
\begin{subfigure}[t]{0.24\textwidth}
    \centering
    \includegraphics[width=\linewidth]{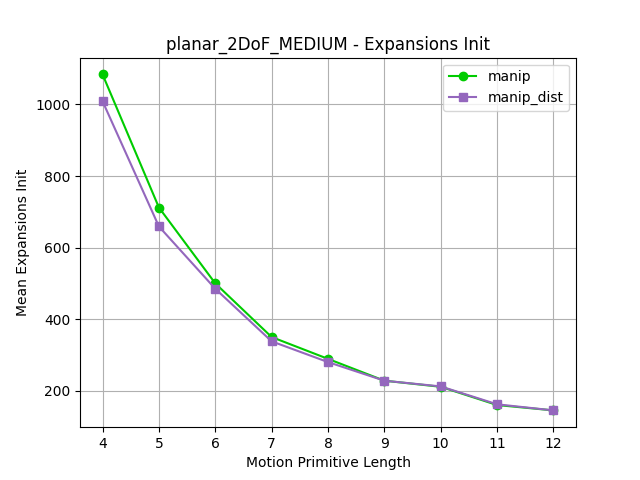}
    \caption{\texttt{2DoF\_MEDIUM} – Expansions}
\end{subfigure}
\hfill
\begin{subfigure}[t]{0.24\textwidth}
    \centering
    \includegraphics[width=\linewidth]{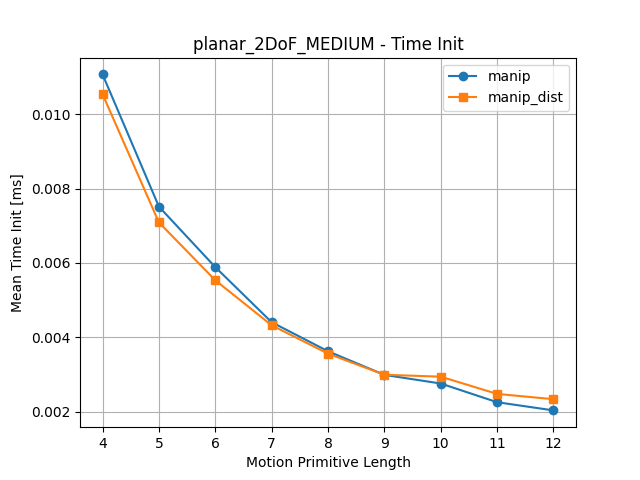}
    \caption{\texttt{2DoF\_MEDIUM} – Planning Time}
\end{subfigure}

\vspace{0.5em}

\begin{subfigure}[t]{0.24\textwidth}
    \centering
    \includegraphics[width=\linewidth]{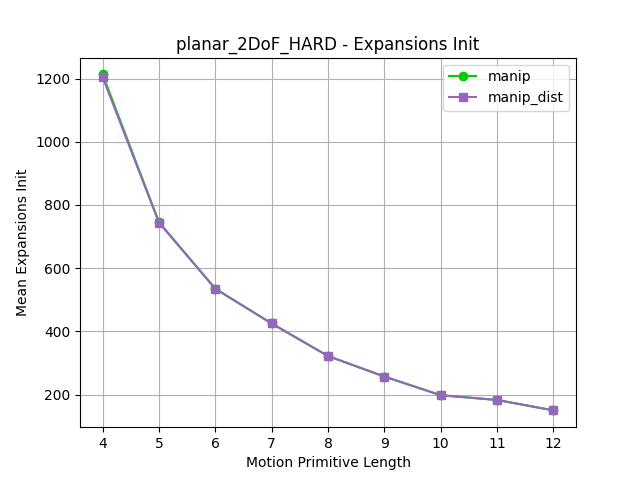}
    \caption{\texttt{2DoF\_HARD} – Expansions}
\end{subfigure}
\hfill
\begin{subfigure}[t]{0.24\textwidth}
    \centering
    \includegraphics[width=\linewidth]{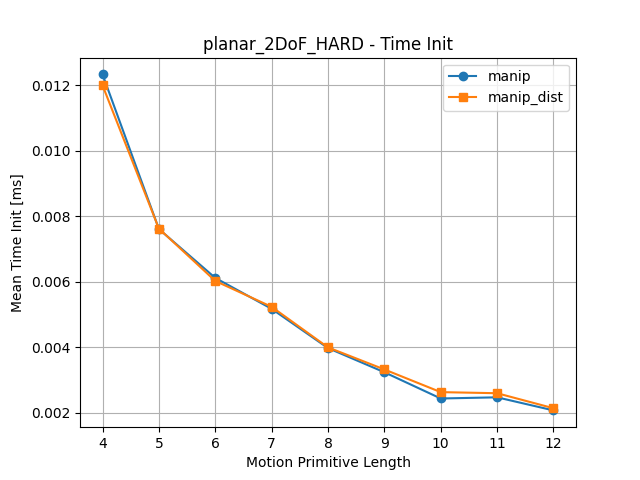}
    \caption{\texttt{2DoF\_HARD} – Planning Time}
\end{subfigure}
\hfill
\begin{subfigure}[t]{0.24\textwidth}
    \centering
    \includegraphics[width=\linewidth]{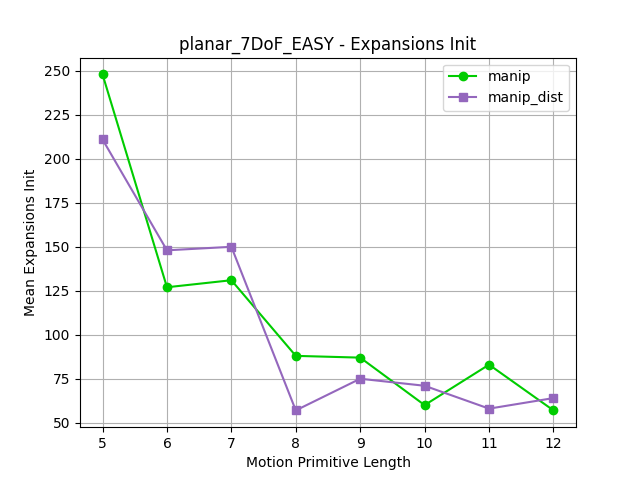}
    \caption{\texttt{7DoF\_EASY} – Expansions}
\end{subfigure}
\hfill
\begin{subfigure}[t]{0.24\textwidth}
    \centering
    \includegraphics[width=\linewidth]{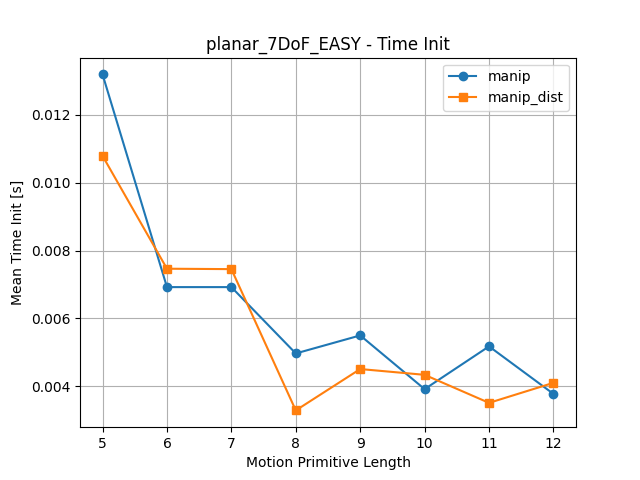}
    \caption{\texttt{7DoF\_EASY} – Planning Time}
\end{subfigure}

\vspace{0.5em}

\begin{subfigure}[t]{0.24\textwidth}
    \centering
    \includegraphics[width=\linewidth]{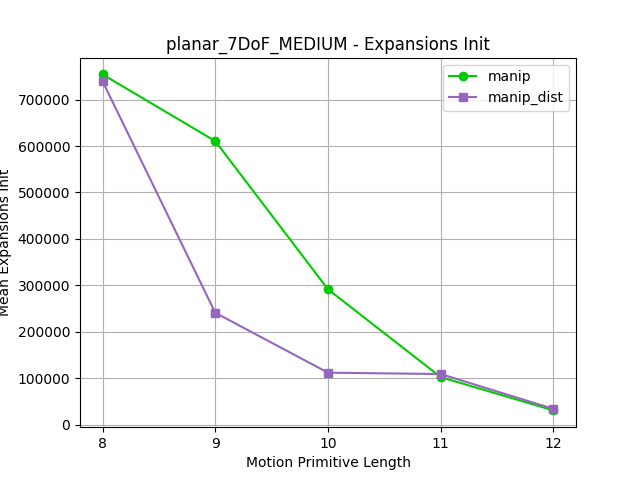}
    \caption{\texttt{7DoF\_MEDIUM} – Expansions}
\end{subfigure}
\hfill
\begin{subfigure}[t]{0.24\textwidth}
    \centering
    \includegraphics[width=\linewidth]{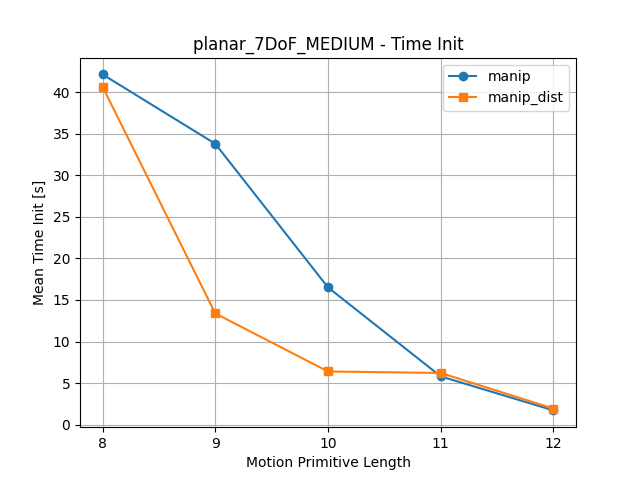}
    \caption{\texttt{7DoF\_MEDIUM} – Planning Time}
\end{subfigure}
\hfill
\begin{subfigure}[t]{0.24\textwidth}
    \centering
    \includegraphics[width=\linewidth]{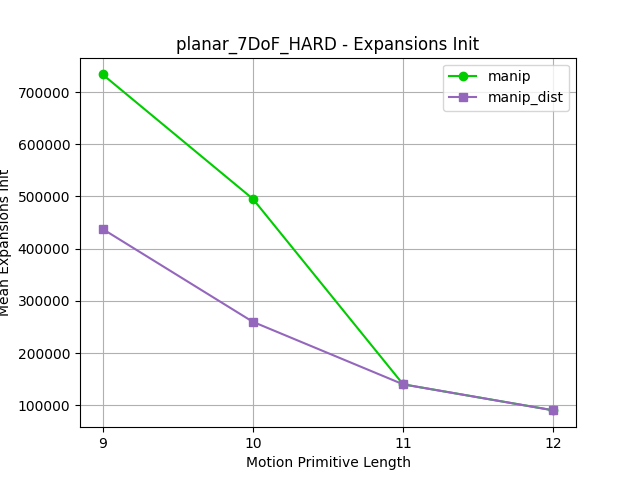}
    \caption{\texttt{7DoF\_HARD} – Expansions}
\end{subfigure}
\hfill
\begin{subfigure}[t]{0.24\textwidth}
    \centering
    \includegraphics[width=\linewidth]{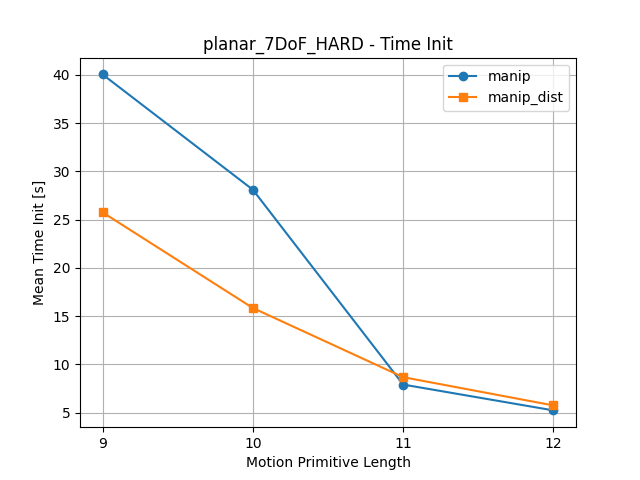}
    \caption{\texttt{7DoF\_HARD} – Planning Time}
    \label{fig:zadnjiGrafik}
\end{subfigure}

\caption{Comparison of initial planning time and number of expansions for fixed motion primitives and burs across all planning scenarios.}
\label{fig:all_scenarios}
\end{figure*}

The experimental results are presented in Tables~\ref{prvatabela}--\ref{zadnjatabela}. In addition, Figures~\ref{fig:prviGrafik}--\ref{fig:zadnjiGrafik} illustrate the initial planning time and the number of expansions, as measures of algorithm efficiency in finding any solution—not necessarily optimal. For the seven-joint manipulator, no algorithm found an optimal solution within the allocated planning and repair time, making it impossible to compare solution quality. Therefore, only initial planning time and initial number of expansions are shown in the figures \ref{fig:prviGrafik}-\ref{fig:zadnjiGrafik}. 
\\ \indent The proposed algorithm using burs of free configuration space generally outperforms the competing algorithm which relies on fixed-length motion primitives. This advantage is attributed to its ability to more efficiently explore areas where the minimum distance between the robot and obstacles is large. The algorithm shows particularly good results in complex, high-dimensional scenarios. In higher-dimensional planning scenarios, the bur-based algorithm finds a solution up to 60\% faster and reduces the number of expansions by as much as 60\%. 
The fixed-primitive algorithm performs slightly better at coarser resolutions (i.e., larger primitive lengths), as the bur spines often cannot exceed the primitive length. In these cases, computing the minimum distance and performing collision checks adds overhead, whereas the fixed-primitive method only performs collision checking. Consequently, the use of burs becomes suboptimal due to frequent computation of minimum distances, which is the most computationally expensive part of the algorithm. Also, the fixed-primitive algorithm performs marginally better in terms of planning times in the complex low-dimensionality scenario. As for 2-DoF scenarios, both algorithms succeed in finding optimal solutions. 
\\ \indent Analyzing the relationship between planning time and number of expansions versus primitive length, we observe that the bur-based graph search algorithm is significantly less sensitive to the resolution of the search graph.

\section{Conclusion}
This paper proposed a motion planning algorithm for robotic manipulators that combines sampling-based and search-based methods to improve efficiency in complex planning scenarios. The core contribution lies in introducing burs of free configuration space as adaptive motion primitives for generating successor states during graph search.
\\ \indent The comparative simulation study revealed that the proposed approach yields shorter planning times and fewer node expansions in most scenarios, compared to an approach using fixed-length motion primitives. The advantage was especially notable in more complex environments involving manipulators with higher degrees of freedom. In simpler scenarios or when using coarser resolutions, both approaches produced comparable results, indicating that the choice of the method is up to the specific planning problem. \\
\indent Future work will focus on integrating generalized burs with search-based planning methods. Additionally, it would be beneficial to develop a more efficient robot representation that allows computing $d_c$ using a smaller number of collision spheres. Since bur dimensions depend solely on the robot’s minimum distance to obstacles, the resulting neighbor nodes may be far from the optimal path. This may require generating additional intermediate nodes to find optimal solutions. A potential research direction is to investigate efficient placement of such intermediate nodes along bur spines—not just at their endpoints. Moreover, future work should explore the development of a heuristic function that better fits the proposed algorithm.

\end{document}